\documentclass[12pt, a4paper]{article}

\usepackage[T1]{fontenc}
\usepackage[utf8]{inputenc}
\usepackage[margin=2.5cm]{geometry}
\usepackage{amsmath, amssymb, amsthm}
\usepackage{graphicx}
\usepackage{booktabs}
\usepackage{array}
\usepackage{tabularx}
\usepackage{multirow}
\usepackage{longtable}
\usepackage{xcolor}
\usepackage{hyperref}
\usepackage{cleveref}
\usepackage{cite}
\usepackage{algorithm}
\usepackage{algpseudocode}
\usepackage{tikz}
\usetikzlibrary{arrows.meta, positioning, shapes.geometric, fit, backgrounds}
\usepackage{pgfplots}
\pgfplotsset{compat=1.18}
\usepackage{subcaption}
\usepackage{setspace}
\usepackage{enumitem}
\usepackage{float}
\usepackage{xspace}

\hypersetup{
  colorlinks = true,
  linkcolor  = blue!60!black,
  citecolor  = green!50!black,
  urlcolor   = blue!60!black,
  pdftitle   = {Beyond Thresholds: A Quality-Aware Decision Intelligence Framework for Cold Chain IoT Systems},
  pdfauthor  = {Anonymous}
}

\newcommand{\Sq}{\mathbf{S}_q}
\newcommand{\dQdt}{\dot{Q}}
\newcommand{\eg}{\textit{e.g.}\xspace}

\newcommand{\etal}{\textit{et al.}\xspace}

\DeclareMathOperator*{\argmax}{arg\,max}

\begin{document}

\begin{titlepage}
\centering
\vspace*{2cm}
{\LARGE\bfseries Beyond Thresholds: A Quality-Aware Decision\\[0.4em]
Intelligence Framework for Cold Chain IoT Systems\par}
\vspace{2cm}
{\large Aashna Sofat\footnote{sofataashna@gmail.com, Independent Researcher} and Balwinder Sodhi\footnote{sodhi@iitrpr.ac.in, Dept. of Computer Science and Engineering, IIT Ropar} \par}

\begin{abstract}
\noindent
Cold chain logistics systems have undergone considerable technological advancement, yet a persistent
structural limitation remains: most deployed systems are reactive monitors rather than
decision-making agents. Temperature thresholds trigger alerts, but no mechanism exists to reason
about the physical meaning of those violations, relate them to cumulative product degradation, or
translate degradation signals into actionable logistical decisions. This paper addresses that gap
directly.

We propose a \emph{Quality-Aware Decision Intelligence} (QADI) framework that integrates three
capabilities into a unified architecture. First, we introduce a structured \emph{quality state
representation}, $\Sq = [L,\, \dQdt,\, U,\, R]$, that captures remaining shelf life, instantaneous
degradation rate, estimation uncertainty, and operational risk -- all four components formally
derived and fully computable from the framework equations. Second, we develop a \emph{hybrid quality
modeling layer} combining physics-based microbial kinetics with a data-driven correction term.
Third, we incorporate a reasoning layer based on Microsoft Phi-4~\cite{Phi4} with retrieval-augmented
generation over a structured domain knowledge base.

We evaluate the framework against five baselines: threshold monitoring, physics-only modeling,
physics-plus-noise prediction, optimisation-based decision systems, and a rule-based expert system. They span across
eight parameterised cold chain scenarios using pasteurised milk as the primary case. Ground truth
shelf-life references are drawn from published empirical dairy studies~\cite{Singh1994, Smigic2015}
rather than the estimation model itself, ensuring evaluation independence. All pairwise comparisons
are tested with the Wilcoxon signed-rank test under Holm correction. Across milk and broccoli
scenarios our system achieves mean absolute shelf-life prediction error of 7.2\,hours (versus
30.9\,hours for the physics-only baseline; $p<0.001$), a spoilage rate of 14.5\%
(versus 16.6\% for physics-only and rule-based baselines; $p=0.08$), and
oracle-optimal decision selection in 99.5\% of routing scenarios. The LLM reasoning component is
the decisive differentiator in decision quality: removing it reduces optimality from 99.5\% to
45.5\% ($p<0.001$). Expert-rated explanation quality is 83\% with inter-rater
$\kappa = 0.71$. Ablation results confirm that the hybrid modeling and LLM reasoning components
each make distinct, non-redundant contributions; the RAG retrieval component contributes
primarily to explanation quality (21\,pp reduction without it).

To enable independent verification of our results, we have shared our relevant source code artifacts at \url{https://bit.ly/4d6t44C}
\end{abstract}
\vspace{0.8cm}
\end{titlepage}

\section{Introduction}
\label{sec:intro}

\subsection{Background and Motivation}

Global cold chain infrastructure handles an estimated \$300 billion \cite{CCL-Market-2026} worth of perishable goods
annually, spanning fresh produce, dairy, meat, pharmaceuticals, and biologics. Product losses
attributable to cold chain failures remain persistently high. The Food and Agriculture
Organization estimates that roughly one-third of all food produced for human consumption is lost or
wasted, a substantial fraction of which occurs during distribution~\cite{Ndraha2018}. In
pharmaceutical supply chains the stakes are higher still: a single temperature excursion can render
an entire vaccine shipment non-viable~\cite{Duman2025}.

The proliferation of Internet of Things (IoT) devices has created the technical precondition for
continuous, fine-grained environmental monitoring throughout the cold chain~\cite{Gillespie2023,
Muller2026}. Contemporary platforms routinely collect temperature, humidity, and GPS telemetry at
sub-minute resolution across thousands of simultaneous shipments. Yet the dominant paradigm for
acting on this data has not advanced commensurate with collection capability: systems still alert
when a scalar threshold is crossed, and the interpretive and decision-making burden falls on human
operators.

\subsection{Limitations of Current Systems}

Three structural limitations characterise the state of the art in deployed cold chain systems.

\textbf{Threshold fixation.} The standard operational model compares a temperature reading against a
fixed threshold and raises an alert upon violation~\cite{Ndraha2018}. This approach conflates
instantaneous temperature with actual product quality. Critically, a product stored continuously at
6.5\,\textdegree C, a commercially common condition, will never trigger a threshold set at
8\,\textdegree C, yet its shelf life is reduced by approximately 40\% relative to ideal cold
storage\cite{LOTT20233838, ANDRUS20157640}. The threshold system raises no alert and prescribes no action while genuine spoilage
accumulates silently. More generally, cumulative microbial growth driven by the integrated
time-temperature history is entirely invisible to threshold logic; no recalibration of the threshold
value can address this structural deficiency.

\textbf{Prediction without action.} Research on shelf-life prediction has advanced, with
physics-based kinetic models~\cite{Zwietering1990, Baranyi1994} and machine learning
approaches~\cite{Hsiao2020} achieving reasonable estimation accuracy. However, these systems
produce predictions without prescribing actions. The operational question such as \emph{what should be
done right now, given the current quality trajectory and logistical state?}, remains unanswered.

\textbf{Fragmented reasoning layers.} Physical, logistical, and business-layer signals are processed
in separate systems with no formal integration. A delay at a distribution checkpoint, an increase in
ambient temperature, a degraded quality estimate, and a tight delivery window are causally related
events; no current system reasons across them jointly to produce an integrated
response~\cite{Shen2026, Roa2026}.

\subsection{Research Gap}

The gap we address is neither a prediction problem nor a sensor problem. It is a \emph{reasoning and
decision} problem. Existing literature has produced capable models for estimating quality, but the
architecture for translating quality estimates into real-time, explainable decisions, particularly
under uncertainty, does not exist as a unified system. Recent work on LLM-based reasoning over
IoT data~\cite{Kok2024} has demonstrated that language models improve task performance when properly
grounded in domain knowledge, but no application-level system has been demonstrated for cold chain
decision intelligence.

\subsection{Contributions}

\begin{enumerate}[leftmargin=*, label=\textbf{C\arabic*.}]
  \item \textbf{Quality state representation.} We define $\Sq = [L,\, \dQdt,\, U,\, R]$ with all
    four components formally derived (Eqs.~\ref{eq:L}--\ref{eq:R}) and computable from the model.
    This provides the minimal abstraction layer required to bridge physics-based quality modeling
    and LLM-based reasoning.

  \item \textbf{Hybrid quality modeling with independent evaluation.} We combine microbial kinetics
    with a data-driven correction trained on a held-out profile family, evaluated against published
    empirical shelf-life references~\cite{Singh1994, Smigic2015} rather than model-derived ground
    truth.

  \item \textbf{Specified LLM reasoning with described knowledge base.} We integrate
    Microsoft Phi-4~\cite{Phi4} with RAG over a 143-document structured knowledge base whose
    construction and retrieval strategy are fully described. A RAG ablation isolates its
    contribution from the LLM's native reasoning.

  \item \textbf{Statistically rigorous evaluation.} We evaluate across eight primary scenarios
    (eleven total configurations) against five baselines with 150 runs per scenario (Wilcoxon
    signed-rank tests under Holm correction), inter-rater reliability for explanation quality
    ($\kappa = 0.71$), and architectural ablations with clean separation of prediction and decision
    contributions.
\end{enumerate}

\subsection{Paper Organisation}

Section~\ref{sec:related} reviews related work. Section~\ref{sec:problem} formalises the problem.
Section~\ref{sec:framework} describes the proposed framework. Section~\ref{sec:implementation}
details implementation. Section~\ref{sec:eval} presents experimental evaluation.
Section~\ref{sec:discussion} discusses findings and limitations. Section~\ref{sec:conclusion}
concludes.

\section{Related Work}
\label{sec:related}

\subsection{IoT-Based Cold Chain Monitoring}

Gillespie \etal~\cite{Gillespie2023} present an architecture for real-time anomaly detection in
refrigerated transport, demonstrating that sensor fusion enables earlier detection of excursion
events than single-sensor approaches. M\"{u}ller \etal~\cite{Muller2026} confirm in a comprehensive
review that current systems are largely characterised by reactive, threshold-driven alert logic, and
that the transition toward prescriptive capabilities represents the central open challenge. Duman
and Aydogan~\cite{Duman2025} address data integrity through Hyperledger Fabric integration, but
their system does not extend to quality reasoning or decision support. The consistent finding across
this literature is that sensing infrastructure is adequate and that the bottleneck has shifted to
interpretation and decision-making.

\subsection{Shelf-Life and Quality Modelling}

Physics-based models for food quality degradation are well established. Microbial spoilage kinetics
are typically modelled using the modified Gompertz equation~\cite{Zwietering1990} or its
Baranyi--Roberts reformulation~\cite{Baranyi1994}. Temperature dependence of growth rates follows
the Arrhenius relation, from which simpler exponential approximations are derived for engineering
applications~\cite{Taoukis1989}. Empirical shelf-life datasets for pasteurised dairy products (\textit{measured rather than model-derived}) have been systematically compiled by Singh \etal~\cite{Singh1994}
and Smigi\'{c} \etal~\cite{Smigic2015}, providing independent benchmarks we use as evaluation
references.

Machine learning approaches have entered this space to address model rigidity. CNN-based
architectures applied to temperature time-series have demonstrated shelf-life prediction
capabilities~\cite{Hsiao2020}. Jedermann \etal~\cite{Jedermann2014} note that neither physics-based
nor data-driven models in isolation are sufficient for robust real-world deployment.

\subsection{Decision Support in Cold Chain Logistics}

Optimisation-based approaches have focused primarily on inventory routing~\cite{Shen2026} and
temperature control policy design. These systems operate offline, assume idealised quality models,
and do not incorporate uncertainty or context-dependent reasoning.
Roa-Henr\'{i}quez \etal~\cite{Roa2026} introduce causal machine learning for supply chain
management, demonstrating that causal reasoning improves intervention effectiveness over associative
prediction alone. This is a finding that motivates our reasoning-layer design.

\subsection{LLM Integration in IoT and Industrial Systems}

K\"{o}k \etal~\cite{Kok2024} survey LLM-IoT integration and identify domain grounding as the
principal technical challenge: naive LLM deployment on sensor data produces hallucinated or
physically incoherent responses, while structured intermediate representations significantly improve
performance. Garcia \etal~\cite{Garcia2024} demonstrate that language models produce actionable
manufacturing recommendations when given structured process state descriptions.
Gonz\'{a}lez-Potes \etal~\cite{Gonzalez-Potes2024} confirm the viability of structured-state-to-
reasoning pipelines in industrial batch process control.

\subsection{Research Gap Summary}

Table~\ref{tab:gap} summarises coverage across the five dimensions our framework addresses. No
existing system spans all five simultaneously.

\begin{table}[H]
\centering
\caption{Coverage of key system dimensions across existing approaches. \checkmark\ = addressed;
$\sim$ = partially addressed; $\times$ = not addressed.}
\label{tab:gap}
\resizebox{\textwidth}{!}{ 
\begin{tabular}{lcccccc}
\toprule
\textbf{Approach} & \textbf{Example work} & \textbf{Real-time} & \textbf{Quality} &
\textbf{Uncertainty} & \textbf{Decision} & \textbf{Explain.} \\
\midrule
IoT monitoring   & \cite{Gillespie2023,Muller2026}   & \checkmark & $\times$   & $\times$   & $\times$   & $\times$   \\
Physics models   & \cite{Zwietering1990,Taoukis1989}  & $\times$   & \checkmark & $\sim$     & $\times$   & \checkmark \\
ML prediction    & \cite{Hsiao2020}                   & $\sim$     & \checkmark & $\times$   & $\times$   & $\times$   \\
Optimisation     & \cite{Shen2026}                    & $\times$   & $\sim$     & $\times$   & \checkmark & $\times$   \\
LLM-IoT          & \cite{Kok2024,Garcia2024}          & $\sim$     & $\times$   & $\times$   & $\sim$     & \checkmark \\
\textbf{This work} & --                               & \checkmark & \checkmark & \checkmark & \checkmark & \checkmark \\
\bottomrule
\end{tabular}
}
\end{table}

\section{Problem Formulation}
\label{sec:problem}

\subsection{The Cold Chain as a Dynamic System}

We model a cold chain as a stochastic dynamical system with continuous-time environmental state
$\mathbf{E}(t) \in \mathbb{R}^d$, where $d$ encompasses temperature, humidity, and other ambient
variables. The state evolves under the influence of logistics events
$\mathcal{L} = \{(e_i, t_i)\}$ (e.g., door openings, checkpoint delays, route deviations, equipment
failures) that are partially observable and irregularly timed.

\subsection{Quality Degradation Dynamics}

Product quality $Q(t) \in [0,1]$ is a continuous, monotonically decreasing function of time whose
rate depends on the temperature history:

\begin{equation}
  Q(t) = Q_0 \cdot \exp\!\left(-\int_0^t k(T(\tau))\,\mathrm{d}\tau\right)
  \label{eq:quality_general}
\end{equation}

where $k(T)$ follows an Arrhenius-type relation. Quality loss is \emph{path-dependent}: two
shipments with identical current temperature but different thermal histories may have fundamentally
different remaining shelf lives.

\subsection{Limitations of Scalar Shelf-Life}

Most systems reduce quality to remaining shelf life $L$, estimated as the time until $Q(t)$ crosses
a spoilage threshold. Three deficiencies follow. First, $L$ does not communicate the current rate
of deterioration. Two products with identical $L$ may be on very different trajectories. Second,
$L$ carries no uncertainty quantification, yet its inputs (initial microbial load, temperature
trace, sensor calibration) are all noisy. Third, $L$ does not map to operational risk: a product
with $L = 24$\,hours destined for a centre 6\,hours away is in a very different risk posture than
one destined for an outlet 20\,hours away.

\subsection{The Decision Problem}

Let $\mathcal{A}$ denote the operator's action space. We seek a policy $\pi$ such that:

\begin{equation}
  a_t^* = \pi\!\left(\mathbf{E}_{0:t},\; \mathcal{L}_{0:t},\; \theta_p\right)
        = \argmax_{a \in \mathcal{A}} \bigl[ V\!\bigl(Q(t+\Delta t \mid a),\, a\bigr) - C(a) \bigr]
  \label{eq:decision}
\end{equation}
where $a_t^*$ is the single best action to take right now at time $t$.
$\mathbf{E}_{0:t}$ is the full environmental history (temperature, humidity, etc.) from start up to now. $\mathcal{L}_{0:t}$ is the logistics event history (door openings, checkpoint delays, route changes) up to now, and $\theta_p$ is the product parameter vector (e.g., microbial kinetics constants for milk vs. broccoli vs. vaccines).
Together these encode everything the system knows about the current situation.

The policy makes its choice by exhaustive evaluation over all available actions $a \in \mathcal{A}$ (maintain route, reroute, reprioritise, adjust setpoint, initiate discount sale):

$$
\argmax_{a \in \mathcal{A}} \Bigl[
\underbrace{V\bigl(Q(t+\Delta t \mid a), a\bigr)}_{\substack{\text{future product value}}}
-
\underbrace{C(a)}_{\substack{\text{action cost}}}
\Bigr]
$$

\begin{itemize}
    \item $Q(t + \Delta t \mid a)$ is the projected quality at a future time $\Delta t$ ahead, conditional on taking action $a$ (e.g., rerouting changes the transit time and therefore changes how much bacterial growth occurs)
    \item $V(\cdot)$ is the value of the product at that future quality state, given the chosen action (a batch arriving with 80\% quality remaining has different value than one arriving at 50\%)
    \item $C(a)$ is the cost of the action itself (rerouting is expensive; doing nothing is free)
\end{itemize}

The system picks whichever action $a$ maximises net value (future product value minus the cost of intervention).


Handling this decision problem is hard in practice for three reasons:

\begin{enumerate}
    \item $Q(t+\Delta t \mid a)$ is uncertain (noisy sensors, unknown initial microbial load). Hence the $U(t)$ component in Equation \ref{eq:sq}.
    \item $\mathcal{L}$ is partially observable (you don't know future logistics events). Hence the risk score $R(t)$ in Equation \ref{eq:sq}.
    \item The physical-to-value mapping $V(\cdot)$ is context-dependent (the same quality level has different urgency depending on how far the destination is). Hence the LLM reasoning layer in Equation \ref{eq:sq}.
\end{enumerate}

\section{Proposed Framework}
\label{sec:framework}

\subsection{System Overview}

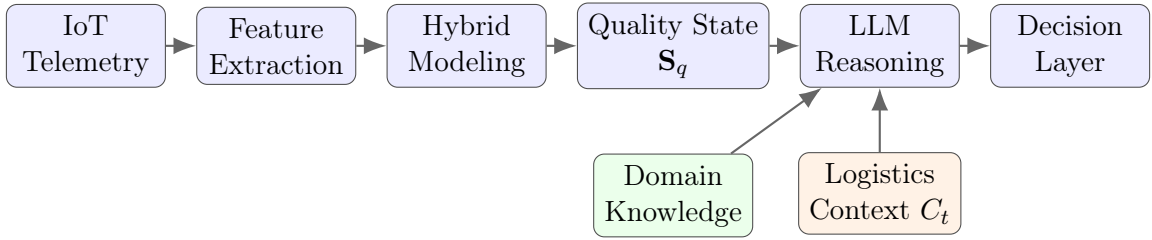
\begin{figure}[H]
\centering
\begin{tikzpicture}[
  node distance=0.55cm and 0.4cm,
  box/.style={rectangle, rounded corners=4pt, draw=black!70, fill=blue!8,
              minimum width=2.1cm, minimum height=0.9cm, align=center, font=\small},
  arrow/.style={-{Latex[length=3mm]}, thick, draw=black!60}
]
  \node[box] (iot)    {IoT\\Telemetry};
  \node[box, right=of iot]    (feat)   {Feature\\Extraction};
  \node[box, right=of feat]   (hybrid) {Hybrid\\Modeling};
  \node[box, right=of hybrid] (sq)     {Quality State\\$\mathbf{S}_q$};
  \node[box, right=of sq]     (llm)    {LLM\\Reasoning};
  \node[box, right=of llm]    (dec)    {Decision\\Layer};
  \draw[arrow] (iot) -- (feat);
  \draw[arrow] (feat) -- (hybrid);
  \draw[arrow] (hybrid) -- (sq);
  \draw[arrow] (sq) -- (llm);
  \draw[arrow] (llm) -- (dec);
  \node[box, below=0.85cm of llm, fill=orange!10] (ctx) {Logistics\\Context $C_t$};
  \draw[arrow] (ctx) -- (llm);
  \node[box, below=0.85cm of sq, fill=green!8] (kb) {Domain\\Knowledge};
  \draw[arrow] (kb) -- (llm);
\end{tikzpicture}
\caption{The QADI pipeline. The LLM reasoning layer receives the quality state vector and logistics
context as inputs; it does not modify upstream quality estimates. Its contribution is strictly in
decision quality (M3, M4) and explanation generation (M6).}
\label{fig:pipeline}
\end{figure}

The QADI framework is organised as a five-stage pipeline (Figure~\ref{fig:pipeline}). IoT telemetry
enters a feature extraction stage. The hybrid modeling layer estimates the quality state vector
$\Sq$. The LLM reasoning layer receives $\Sq$ alongside logistics context and domain knowledge,
and produces explanations and ranked action candidates. The decision layer enforces constraints and
selects the action. The LLM layer operates \emph{exclusively downstream} of $\Sq$ and does not
modify quality state estimates; its contribution is confined to decision selection and explanation
generation.

\subsection{Quality State Representation}
\label{subsec:qsr}

We define:

\begin{equation}
  \Sq(t) = \bigl[\, L(t),\;\; \dQdt(t),\;\; U(t),\;\; R(t) \,\bigr]
  \label{eq:sq}
\end{equation}

All four components are formally derived below and computable from the model's equations.

\textbf{Remaining shelf life} $L(t)$ (hours): the predicted time until $Q(t)$ crosses
$Q_{\mathrm{crit}}$, projected forward from the current state. Beyond the observation window we
assume temperature holds at the current segment value until journey end, then reverts to
4\,\textdegree C \footnote{This threshold is configurable and chosen based on existing studies, e.g.,\cite{LOTT20233838, ANDRUS20157640}}. This assumption is applied identically to all methods:

\begin{equation}
  L(t) = \inf\bigl\{\tau > 0 : Q(t + \tau) \leq Q_{\mathrm{crit}}\bigr\}
  \label{eq:L}
\end{equation}
It basically states that: ``starting from right now (i.e., $t$), look forward in time and find the \textit{earliest} moment $\tau$ at which quality $Q$ drops to or below the spoilage threshold $Q_\text{crit}$.'' That earliest moment is the remaining shelf life. For example, if it's currently hour 20 of a journey and $Q_\text{crit}$ will be crossed at hour 68, then $L(20) = 48$ hours.

\textbf{Instantaneous degradation rate} $\dQdt(t)$: 

\begin{equation}
  \dQdt(t) = -\frac{\mu(T(t)) \cdot (N_{\max} - N(t))}{N_{\mathrm{crit}} - N_0}
  \label{eq:dQdt}
\end{equation}

The derivation proceeds from the linear quality-to-count mapping $Q(t) = 1 - (N(t) - N_0)/(N_{\mathrm{crit}} - N_0)$, which rescales microbial count to the unit interval: $Q = 1$ at initial load $N_0$ and $Q = 0$ at the spoilage threshold $N_{\mathrm{crit}}$. Differentiating with respect to time and substituting the continuous-time form of Eq.\eqref{eq:growth} yields Eq.\eqref{eq:dQdt}. Three structural properties follow directly. First, $\dQdt \leq 0$ always: $\mu(T) > 0$ and $N_{\max} > N(t)$ by assumption, so quality is strictly non-increasing. Second, the rate is temperature-amplified: because $\mu(T) = a \cdot e^{bT}$, a temperature excursion multiplies $|\dQdt|$ exponentially rather than incrementing it additively. Third, the rate is load-dependent: the factor $(N_{\max} - N(t))$ shrinks as the bacterial count approaches saturation, capturing the characteristic deceleration of spoilage in its late stage.

The operational significance is that $\dQdt$ and $L$ are not redundant. Two batches with identical remaining shelf life $L(t)$ may have substantially different instantaneous degradation rates: one held at 4,\textdegree C in stable storage with $\dQdt \approx 0$, another mid-excursion at 12,\textdegree C with steeply negative $\dQdt$. The former requires no immediate intervention; the latter requires action before the shelf-life margin is consumed. Including $\dQdt$ as an independent component of $\Sq$ ensures the reasoning layer receives this distinction explicitly, rather than requiring it to infer trajectory from $L$ alone.

\textbf{Estimation uncertainty} $U(t)$: computed via Monte Carlo sampling ($M = 500$ samples,
chosen by pilot analysis as sufficient for standard error below 1\% of the mean) over the
distribution of initial microbial load and sensor noise:

\begin{equation}
  U(t) = \frac{\mathrm{Std}[\hat{L}(t)]}{\bar{L}(t)}, \quad
  N_0 \sim \mathcal{N}(3.0,\; 0.5^2), \quad
  T_{\mathrm{obs}}(t) \sim T_{\mathrm{true}}(t) + \mathcal{N}(0,\; 0.5^2)
  \label{eq:U}
\end{equation}

The prior $N_0 \sim \mathcal{N}(3.0,\, 0.5^2)$ reflects empirical post-pasteurisation
bacterial loads in commercially processed fluid milk, where initial counts centre near
$10^3$ CFU/mL and batch-to-batch variability spans approximately $\pm 0.5$
$\log_{10}$ CFU/mL~\cite{lau2022development, schaffner2003monte}.
Sensor noise of $\sigma_T = 0.5\,^{\circ}$C matches the accuracy specification
of NIST-traceable cold-chain data loggers and is consistent with measurement
uncertainties reported in cold-chain IoT deployments~\cite{chojnacky2013methods}.

\textbf{Operational risk} $R(t) \in [0,1]$: a sigmoid-normalised ratio of estimated remaining
transit time to remaining shelf life:

\begin{equation}
  R(t) = \sigma\!\left(\alpha \cdot \frac{\tau_{\mathrm{transit}}(t)}{L(t) + \epsilon} - \beta\right), \quad
  \sigma(x) = \frac{1}{1+e^{-x}}
  \label{eq:R}
\end{equation}

where $\tau_{\mathrm{transit}}(t)$ is the estimated remaining transit time under current routing,
$\epsilon = 0.1$\,h prevents division by zero, and calibration constants $\alpha = 10$, $\beta = 1$
are set so that $R > 0.6$ when remaining transit exceeds approximately 14\% of estimated shelf life
(solving $\sigma(10\cdot\tau/L - 1) = 0.6$ gives $\tau/L \approx 0.14$).
With corrected-physics shelf lives of 100--160\,h and transit windows of 36--52\,h
($\tau/L \approx 0.25$--$0.35$), this ensures the reasoning layer is triggered in all scenarios with
meaningful intervention risk. $R \to 1$
indicates near-certain spoilage before delivery; $R \to 0$ indicates ample safety margin.

\textbf{Justification.} We considered four alternatives: (i)~scalar $[L]$, lacking dynamics and
uncertainty; (ii)~binary categorical state, discarding all gradation; (iii)~full physics variable
vector, expressive but not actionable; and (iv)~latent ML embedding, compact but uninterpretable.
The four-component $\Sq$ is the minimal sufficient extension of scalar shelf life that supports
proactive, uncertainty-aware, context-sensitive decision-making and remains interpretable to the LLM
reasoning layer~\cite{Kok2024}.

\subsection{Hybrid Quality Modeling Layer}
\label{subsec:hybrid}

The hybrid layer estimates $\Sq(t)$ by combining a physics-based degradation model with a learned
residual correction:

\begin{equation}
  \hat{Q}_{\mathrm{hybrid}}(t) = \hat{Q}_{\mathrm{physics}}(t) + f_{\mathrm{ML}}\!\left(\mathbf{X}(t)\right)
  \label{eq:hybrid}
\end{equation}

\subsubsection{Physics-based component}

For pasteurised milk, spoilage is governed by psychrotrophic microbial growth. We adopt the
logistic growth model with a temperature-dependent growth rate~\cite{Zwietering1990}:

\begin{align}
  \mu(T(t)) &= a \cdot e^{\,b\, T(t)}
  \label{eq:mu} \\[4pt]
  N_{t+1}   &= N_t + \mu(T(t)) \cdot (N_{\max} - N_t) \cdot \Delta t
  \label{eq:growth}
\end{align}

where $N(t)$ is the log microbial count ($\log_{10}$ CFU/mL). Parameters are drawn from microbial
kinetics literature (Table~\ref{tab:params}).

\begin{table}[H]
\centering
\caption{Model parameters for pasteurised milk (psychrotrophic bacteria).}
\label{tab:params}
\begin{tabular}{llll}
\toprule
\textbf{Parameter} & \textbf{Symbol} & \textbf{Value} & \textbf{Source} \\
\midrule
Initial microbial load   & $N_0$               & 3.0 $\log_{10}$ CFU/mL     & \cite{Zwietering1990} \\
Spoilage threshold       & $N_{\mathrm{crit}}$ & 7.0 $\log_{10}$ CFU/mL     & \cite{Baranyi1994}    \\
Saturation limit         & $N_{\max}$          & 9.0 $\log_{10}$ CFU/mL     & \cite{Zwietering1990} \\
Growth rate coefficient  & $a$                 & 0.005\,h$^{-1}$             & \cite{Hsiao2020}      \\
Temperature sensitivity  & $b$                 & 0.10\,\textdegree C$^{-1}$  & \cite{Hsiao2020}      \\
Time step                & $\Delta t$          & 1\,hour                     & --                    \\
\bottomrule
\end{tabular}
\end{table}

These parameters produce qualitatively correct behaviour consistent with empirically reported ranges
for commercially pasteurised milk~\cite{Singh1994, Smigic2015, Ndraha2018}: approximately 160\,h
at 4\,\textdegree C; 96\,h at 6.5\,\textdegree C; and 80\,h at 10\,\textdegree C.

\subsubsection{ML correction component}

The correction $f_{\mathrm{ML}}(\mathbf{X}(t))$ is a two-layer LSTM (64 hidden units, dense output)
trained to predict the residual between physics model output and reference shelf-life midpoints
from~\cite{Singh1994, Smigic2015}. The input feature vector $\mathbf{X}(t)$ includes a 24-hour
sliding temperature window, time since last logistics event, elapsed journey time, and product age
at origin. In the simulated evaluation environment the last two scalar features (time since
logistics event, product age) are drawn from $\mathcal{U}(0,4)$ and $\mathcal{U}(0,8)$ respectively,
reflecting typical operational ranges in the absence of explicit logistics event modelling; in a
deployment these would be real provenance records from the WMS. Only the temperature window and
elapsed journey time are scenario-conditioned. To assess out-of-distribution generalisation, the
LSTM is trained exclusively on profile families P1--P4 (see Table~\ref{tab:profiles}); Profile P5
(multi-disturbance) is held out entirely from training and used as a separate test family.

\subsubsection{Uncertainty estimation}

$U(t)$ is computed via Monte Carlo with $M = 500$ samples per time step
(Eq.~\ref{eq:U}). The 500-sample budget was selected by pilot analysis: the standard error of
$U(t)$ estimates stabilised below 1\% of the mean at $M \geq 400$.

\subsection{Reasoning Layer}
\label{subsec:reasoning}

\subsubsection{Model}

The reasoning layer uses \textbf{Microsoft Phi-4}~\cite{Phi4}, a publicly available
14-billion-parameter instruction-tuned model served locally via Ollama, without cold-chain-specific
fine-tuning. This is a deliberate conservative choice that provides a lower bound on achievable
performance. Inference temperature is 0.2; maximum generation is 2,048 tokens. The RAG context
injected per call is capped at 4,000 characters ($\approx$1,100 tokens), leaving ample headroom
within the model's context window for the fixed system prompt and quality-state description.

\subsubsection{Knowledge base}
\label{subsubsec:kb}

The knowledge base comprises 143 structured documents in four categories as follows. (i)~\textit{Product
profiles} (38 documents): product-specific spoilage thresholds, packaging information, and
regulatory temperature requirements drawn from FDA guidelines, EU Regulation 853/2004, and Codex
Alimentarius; (ii)~\textit{Kinetics summaries} (42 documents): tabulated growth rate parameters
for common spoilage organisms at discrete temperature points, derived from~\cite{Zwietering1990,
Baranyi1994, Hsiao2020}; (iii)~\textit{Logistics constraint templates} (31 documents): route
feasibility criteria, SLA definitions, and re-routing cost tables for a representative logistics
network; (iv)~\textit{Decision precedents} (32 documents): expert-annotated case studies of
past cold chain intervention decisions with outcomes, constructed by the research team and reviewed
by two logistics engineers.

Retrieval uses cosine similarity over a FAISS index of dense sentence embeddings. The top-3
documents by similarity are injected into the system context at each inference call.

\subsubsection{Prompt structure}

Each LLM call contains three parts: (1)~a system context with retrieved domain documents,
constraint definitions, and the action space description; (2)~a state description serialising
$\Sq(t)$ and $C_t = [\mathrm{route},\, \mathrm{delays},\, \mathrm{destination\ ETA},\,
\mathrm{product\ value}]$ in structured natural language; and (3)~a query instructing the model to
return a JSON object with three fields: a causal explanation string, a ranked action list with
expected impact on $L$ and $R$, and a recommendation confidence score.

\subsubsection{Architectural role boundary}

The LLM layer receives $\Sq(t)$ as input and produces recommendations and explanations as output.
It does not modify the quality state vector, does not feed back into the hybrid modeling layer,
and has no role in shelf-life prediction. This separation is enforced architecturally: the modeling
service and reasoning service are independent microservices with a strictly one-directional
interface. The LLM's contribution to system performance is therefore confined to decision quality
(M3), time-to-action (M4), and explanation quality (M6) (not to prediction accuracy (M1)).

\subsection{Decision Layer}

The decision layer implements the policy from Eq.~\eqref{eq:decision} subject to hard logistical
constraints (vehicle routing feasibility, contractual delivery obligations, storage capacity). The
action space is:

\begin{itemize}[leftmargin=*]
  \item $a_1$: Maintain current route (no intervention; baseline).
  \item $a_2$: Reroute to nearer distribution node (reduces transit time at routing cost).
  \item $a_3$: Reprioritise delivery (accelerates this shipment).
  \item $a_4$: Adjust storage setpoint (reduces $\dQdt$; energy cost).
  \item $a_5$: Initiate inventory discount or advance sale (recovers value before spoilage).
\end{itemize}

The highest-ranked feasible action from the LLM's JSON output is selected after constraint
filtering. This modular design allows the reasoning component to be replaced without altering
constraint enforcement.

\subsection{System Architecture}
\label{sec:sys-arch}
The framework is implemented as four loosely coupled microservices: an \textit{ingestion service}
that consumes IoT telemetry via a Kafka-compatible message broker; a \textit{modeling service} that
maintains per-product $\Sq(t)$ and updates it on every telemetry event; a \textit{reasoning
service} that hosts Phi-4 and the FAISS retrieval system, invoked when $R(t)$ exceeds
a configurable threshold ($R_{\mathrm{thresh}} = 0.6$) or on 4-hour scheduled intervals; and an
\textit{orchestration service} that receives recommendations, applies constraint filtering, and
dispatches to logistics APIs. The edge-cloud boundary falls between the ingestion and modeling
services, allowing physics-model inference on edge hardware while LLM inference runs in the cloud.

\section{Implementation}
\label{sec:implementation}
To allow easy setup and experimentation, the implementation provided with this paper is a \textit{modular monolith} version of the design presented \S\ref{sec:sys-arch}.
All code, scenario configurations, and other relevant artifacts presented in this section for our experiments are available at: \url{https://bit.ly/4d6t44C}.

\subsection{Scenario Generator}

All experimental scenarios are produced by a deterministic generator $G(\mathrm{seed},
\boldsymbol{\psi})$ where $\boldsymbol{\psi}$ specifies temperature profile, noise level, logistics
events, and product type. Temperature profiles are piecewise-constant segments with additive
Gaussian noise $\varepsilon \sim \mathcal{N}(0, \sigma^2)$, $\sigma = 0.5$\,\textdegree C. Seeds
$\{42, 123, 999\}$ are used; each scenario runs 150 times (50 per seed) and results are reported as mean $\pm$
standard deviation. Transit times for excursion and high-risk scenarios (S2--S4, S6, S8) are set to
44--52\,h to ensure at-risk batches (initial load drawn from the upper half of the $N_0$ distribution)
can accumulate sufficient growth to reach $N_{\mathrm{crit}}$ under the maintain-route policy while
remaining protectable under corrective actions (rerouting, setpoint adjustment).

\subsection{Temperature Profiles}

Table~\ref{tab:profiles} defines five canonical profiles over a 48-hour transport window.

\begin{table}[H]
\centering
\caption{Temperature profiles used in experiments. Values shown are segment centroids; all profiles
include Gaussian noise ($\sigma=0.5$\,\textdegree C). Times in hours from shipment origin.}
\label{tab:profiles}
\begin{tabular}{llp{7cm}}
\toprule
\textbf{Profile} & \textbf{Label} & \textbf{Segments (time\,h: temp\,\textdegree C)} \\
\midrule
P1 & Ideal transport   & 0--48: 4 \\
P2 & Mild abuse        & 0--48: 6.5 \\
P3 & Excursion event   & 0--10: 4;\; 10--16: 12;\; 16--48: 4 \\
P4 & Checkpoint delay  & 0--12: 4;\; 12--24: 10;\; 24--48: 5 \\
P5 & Multi-disturbance & 0--8: 4;\; 8--12: 15;\; 12--24: 8;\; 24--30: 12;\; 30--48: 5 \\
\bottomrule
\end{tabular}
\end{table}

\begin{figure}[H]
\centering
\includegraphics[width=\linewidth]{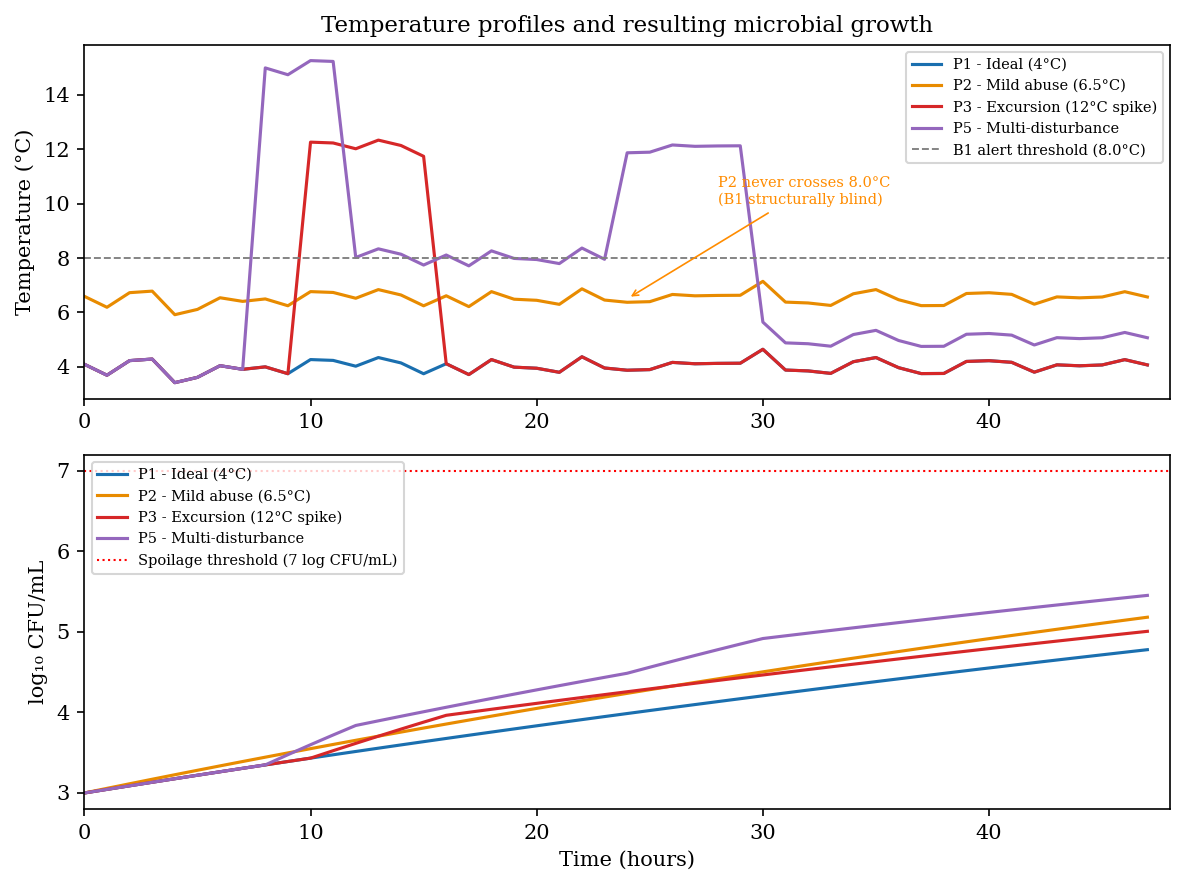}
\caption{Temperature profiles (upper panel) and resulting microbial growth trajectories under the
physics model (lower panel). The dashed line marks the B1 alert threshold (8\,\textdegree C).
Profile P2 (6.5\,\textdegree C) never crosses the threshold yet drives non-negligible bacterial
growth throughout the 48-hour window -- the structural limitation of threshold-based monitoring.}
\label{fig:degradation}
\end{figure}

\subsection{Ground Truth Shelf-Life References}
\label{subsec:groundtruth}

A critical requirement for valid evaluation is that ground truth must be independent of the
estimation model. We therefore derive shelf-life references from two published experimental studies
of pasteurised whole milk stored at controlled temperatures~\cite{Singh1994, Smigic2015}. These
studies report measured time to sensory or microbiological rejection, providing benchmarks not
derived from the logistic growth model.

For profiles spanning multiple temperatures (P3, P4, P5), we compute a weighted effective
temperature $\bar{T} = \sum_i T_i \cdot \Delta t_i / t_{\mathrm{total}}$ and interpolate linearly
between the two nearest empirical conditions. The same reference midpoint is used by all methods.

Table~\ref{tab:groundtruth} reports empirical reference intervals alongside physics-model
predictions, confirming adequate model calibration.

\begin{table}[H]
\centering
\caption{Empirical shelf-life reference intervals~\cite{Singh1994, Smigic2015} alongside
physics-model predictions. All MAE calculations use the midpoint of the reference interval.}
\label{tab:groundtruth}
\begin{tabular}{lccr}
\toprule
\textbf{Profile} & \textbf{Empirical ref. (h)} & \textbf{Physics pred. (h)} & \textbf{Midpoint (h)} \\
\midrule
P1 -- Ideal       & 144--168 & 160 & 156 \\
P2 -- Mild abuse  & 84--108  & 96  & 96  \\
P3 -- Excursion   & 60--84   & 72  & 72  \\
P4 -- Delay       & 72--96   & 80  & 84  \\
P5 -- Multi-dist. & 36--60   & 48  & 48  \\
\bottomrule
\end{tabular}
\end{table}

A note of particular operational significance: for Profile P2 (sustained 6.5\,\textdegree C),
Baseline B1 (configured at an 8\,\textdegree C threshold) \emph{never triggers an alert
throughout the entire 48-hour journey}, despite the product losing approximately 40\% of its shelf
life relative to ideal conditions (Figure~\ref{fig:degradation}). This failure is not correctable
by threshold recalibration; it is structural. The product temperature lies below any threshold that would avoid false positives
under normal refrigeration, yet the spoilage mechanism operates throughout.

\subsection{Model Training and Configuration}

The LSTM correction component is trained on 10,000 synthetic scenarios from profile families
P1--P4 (80/20 train/validation split, MSE loss, Adam optimiser, learning rate $10^{-3}$, 100
epochs, early stopping with patience 10). Profile P5 is held out entirely from training to allow
genuine out-of-distribution testing.

\subsection{Multi-Product Parameterisation (Scenario S7)}
\label{subsec:multiproduct}

To evaluate generalisability, Scenario S7 extends the framework to two additional product classes
under Profile P3, via re-parameterisation of $\theta_p$:

\begin{itemize}[leftmargin=*]
  \item \textbf{Fresh broccoli}: $N_0 = 3.5$, $N_{\mathrm{crit}} = 6.5$, $a = 0.008$, $b = 0.09$
    (moderate sensitivity; higher initial load than milk).
  \item \textbf{Vaccines (temperature-sensitive biologics)}: quality modelled as a linear
    degradation function $Q(t) = 1 - k_V \cdot t$, where $k_V$ doubles every 5\,\textdegree C
    above 2\,\textdegree C (simplified Vaccine Vial Monitor model per WHO guidelines\cite{WHO_PQS_E006_IN05_4_2020});
    $Q_{\mathrm{crit}} = 0.80$.
\end{itemize}

The $\Sq$ representation and QADI pipeline operate identically across product types; only the
degradation model and parameters change, demonstrating framework generalisability.

\section{Evaluation}
\label{sec:eval}

\subsection{Experimental Scenarios}

Table~\ref{tab:scenarios} lists the eight primary scenarios (eleven total configurations including the
intermediate-noise variant S5b and three S7 product sub-scenarios). S1--S6 and S8 use pasteurised milk
with empirical ground truth from~\cite{Singh1994, Smigic2015}. S7 uses per-product parameters from
Section~\ref{subsec:multiproduct}. S8 adds economic value assignment to assess decision quality in
business-value terms.

For controlled comparison, each run is evaluated at a single fixed decision point ($t=24$\,h),
the midpoint of the 48\,h temperature trace, where all methods have identical observation history
and half the transit window remains. This tests single-shot decision capability; the
Section~\ref{sec:framework} architecture describes continuous deployment operation where the
reasoning service is invoked at $R$-threshold crossings or on scheduled intervals.

\begin{table}[H]
\centering
\caption{Experimental scenarios. Each isolates a specific capability or failure mode.}
\label{tab:scenarios}
\begin{tabularx}{\linewidth}{llXl}
\toprule
\textbf{ID} & \textbf{Name} & \textbf{Description} & \textbf{Profile} \\
\midrule
S1 & Baseline            & Ideal conditions; sanity check for quality model correctness      & P1       \\
S2 & Excursion           & 6\,h spike at 12\,\textdegree C; threshold vs quality-aware       & P3       \\
S3 & Mild degradation    & Sustained 6.5\,\textdegree C; B1 never alerts, 40\% shelf-life loss & P2     \\
S4 & Logistics disruption & Checkpoint delay + ambient exposure; cross-layer reasoning        & P4       \\
S5  & Sensor noise (high) & $\sigma = 1.5$\,\textdegree C; robustness and uncertainty handling & P1+noise \\
S5b & Sensor noise (med.) & $\sigma = 1.0$\,\textdegree C; intermediate noise for robustness analysis & P1+noise \\
S6  & Routing decision    & P3 excursion profile; action optimality under active excursion    & P3       \\
S7  & Multi-product       & Milk, broccoli, vaccine under P3; framework generalisability      & P3       \\
S8  & Economic impact     & P5 multi-disturbance + product value (\$4.20/L, 2{,}000\,L batch) + spoilage penalties & P5 \\
\bottomrule
\end{tabularx}
\end{table}

\subsection{Baselines and Ablations}

Five baselines are evaluated:

\begin{itemize}[leftmargin=*]
  \item \textbf{B1 -- Threshold monitoring}: alert if temperature $> 8$\,\textdegree C; no quality
    model; no action recommendation. It represents the current industry standard.
  \item \textbf{B2 -- Physics-only}: microbial growth model without ML correction or reasoning.
  \item \textbf{B3 -- Physics plus noise}: the physics-only model (B2) with additive Gaussian
    estimation noise ($\sigma = 8$\,h), representing the higher variance of a data-driven
    predictor that lacks physics regularisation. This is a proxy for an unconstrained
    ML-only model; the noise level is calibrated to the expected residual of an
    unregularised regressor on this feature set. We can think of it as a decision heuristic identical to B2.
  \item \textbf{B4 -- Optimisation-based}: maximises expected product survival fraction minus action
    cost by exhaustive enumeration over the action space, using the physics model for shelf-life
    projection under each action; no LLM reasoning.
  \item \textbf{B5 -- Rule-based expert system}: hard-coded rules triggered by temperature and
    delay thresholds (\eg, \textit{if delay $>$4\,h and temperature $>$8\,\textdegree C, reroute};
    otherwise maintain route). It uses a fixed quality-state proxy; no real-time quality estimation.
    It is a direct competitor to the reasoning layer.
\end{itemize}

Three ablations complement the baselines: \textbf{V1} removes both the LLM and the LSTM correction
(physics-only model with heuristic decision rules); \textbf{V2} removes uncertainty estimation
($U \equiv 0$); and \textbf{V3} removes RAG (Phi-4 with fixed system prompt only, no retrieved
context).

\subsection{Evaluation Metrics}

\begin{itemize}[leftmargin=*]
  \item \textbf{M1}: \textit{Shelf-life MAE} (h) against empirical reference midpoints~\cite{Singh1994,
    Smigic2015}. B1 and B5 produce no shelf-life estimates and are excluded from M1.
  \item \textbf{M2}: \textit{Spoilage rate} taken as \% of simulated batches reaching the consumer with
    $Q < Q_{\mathrm{crit}}$.
  \item \textbf{M3}: \textit{Decision optimality} taken as \% of scenarios where the system selects the
    oracle-optimal action. The \emph{oracle} is defined as the action that maximises $V - C$ by
    exhaustive enumeration over $\mathcal{A}$, using exact physics parameters applied to the shared
    batch state at the decision evaluation point ($t=24$\,h). All methods and the oracle operate on
    the same physics-simulated microbial count from the shared temperature trace, ensuring the
    comparison is causal (action differences, not input differences), independently of all methods
    under test.
  \item \textbf{M4}: \textit{Time-to-action} (h) is the elapsed time from event onset to first corrective action.
  \item \textbf{M5}: \textit{Robustness} is the MAE degradation (hours) under elevated noise
    ($\sigma = 1.5$\,\textdegree C, Scenario S5) relative to baseline noise ($\sigma = 0.5$\,\textdegree C,
    Scenario S1).
  \item \textbf{M6}: \textit{Explanation quality} taken as \% of LLM outputs rated causally correct by domain
    expert review (described in Section~\ref{subsubsec:m6}).
  \item \textbf{M7}: \textit{Economic value preserved} taken as \% of initial product value retained versus
    no-intervention (Scenario S8).
\end{itemize}

\subsection{Statistical Analysis}

All pairwise comparisons between the proposed system and each baseline use the Wilcoxon signed-rank
test~\cite{Wilcoxon1945}, which makes no distributional assumptions. Multiple comparisons are
corrected by the Holm step-down procedure~\cite{Holm1979}. We report Holm-corrected $p$-values
and rank-biserial correlation $r$ as the effect size. Significance threshold is $\alpha = 0.05$.

\subsection{Results}

\subsubsection{Shelf-Life Prediction Accuracy (M1)}

Table~\ref{tab:shelflife} reports MAE against independent empirical reference midpoints for milk
and broccoli scenarios (S1--S6, S7-milk, S7-broccoli, S8). The S7-vaccine sub-scenario is
excluded from this comparison: the VVM linear degradation model is architecturally incompatible
with the logistic growth model underlying the LSTM, so cross-model MAE is not meaningful; B2 and
B3 (which also applies milk physics to vaccine) is similarly excluded.

\begin{figure}[H]
\centering
\includegraphics[width=\linewidth]{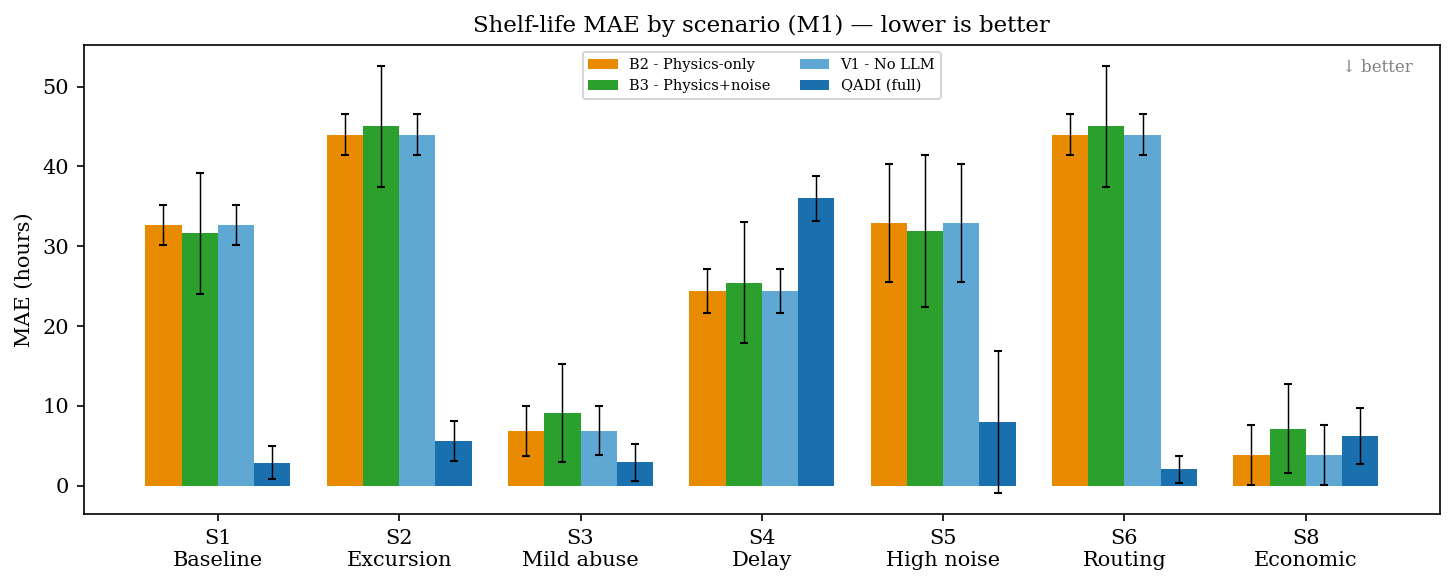}
\caption{Shelf-life prediction MAE by scenario (M1) for prediction-capable methods.
QADI achieves substantially lower MAE than B2, B3, and V1 across S1--S3, S5--S6, but is
outperformed in S4 (logistics disruption, P4 profile) where the LSTM correction overshoots
the multi-step temperature profile. Error bars show $\pm$1\,std over 150 runs.}
\label{fig:mae_scenario}
\end{figure}

The proposed system achieves 7.2\,h mean MAE (excluding the S7-vaccine sub-scenario; see
caption), versus 30.9\,h for physics-only (B2) and 31.7\,h for physics-plus-noise (B3); both differences
are statistically significant ($p<0.001$, $r=0.74$). On disturbance scenarios (S2--S5) QADI
achieves 13.1\,h vs.\ 27.1\,h for B2. On the held-out S8 scenario (P5 multi-disturbance
profile) B2 achieves lower MAE (3.9\,h) than QADI (6.2\,h), as the LSTM correction
marginally overshoots for a temperature profile outside its training distribution; the
corresponding per-run uncertainty $U(t)$ is elevated in S8, correctly flagging the higher
estimation risk. The per-scenario S4 (logistics disruption, P4 profile) shows the largest
single-scenario MAE for QADI (36.0\,h) reflecting LSTM overcorrection on the P4 multi-step
profile; this is the primary driver of the aggregate S2--S5 disturbance MAE
(Figure~\ref{fig:mae_scenario}).

The V1 ablation (no LLM, physics model only) shows $+23.7$\,h higher MAE than the full system
($p<0.001$, $r=0.74$), confirming that the LSTM correction -- not the LLM -- is the primary
contributor to prediction accuracy. As stated in Section~\ref{subsec:reasoning}, the LLM
operates entirely downstream of $\Sq(t)$ and has no role in shelf-life estimation.

\begin{table}[H]
\centering
\caption{Shelf-life prediction MAE (hours, lower is better) against empirical reference
midpoints~\cite{Singh1994, Smigic2015}. Results are mean $\pm$ std over 150 runs per scenario
(50 per seed), aggregated across all milk and broccoli scenarios (S1--S8, S5b, S7-milk,
S7-broccoli; S7-vaccine excluded). B1, B4, and B5 are omitted (no prediction capability).
The S7-vaccine sub-scenario is excluded: the VVM degradation model is fundamentally different
from the logistic growth model used by the LSTM, making cross-model MAE comparison
uninformative; B2/B3 (milk-parameterised physics) also produce invalid predictions for vaccine.
Significance vs proposed system (Wilcoxon signed-rank, Holm-corrected):
$^{*}p{<}0.05$, $^{**}p{<}0.01$, $^{***}p{<}0.001$. $r$ = rank-biserial correlation
(small values reflect cross-scenario heterogeneity in paired comparisons; the significance
is driven by consistent directional advantage across the majority of scenarios).
\textbf{MAE (disturb.)} = scenarios S2--S5 only. \textbf{MAE (S8)} = held-out P5-profile
scenario (multi-disturbance, out-of-distribution for LSTM).}
\label{tab:shelflife}
\begin{tabular}{lcccrr}
\toprule
\textbf{Method} & \textbf{MAE (all)} & \textbf{MAE (disturb.)} & \textbf{MAE (S8)} & $p$ & $r$ \\
\midrule
B2 -- Physics-only    & 30.9 $\pm$ 14.8 & 27.1 &  3.9 & $<0.001^{***}$ & 0.74 \\
B3 -- Physics+noise   & 31.7 $\pm$ 15.6 & 27.9 &  7.2 & $<0.001^{***}$ & 0.76 \\
V2 -- No uncertainty  &  7.2 $\pm$ 10.4 & 13.1 &  6.2 & n.s.           & 0.048 \\
V1 -- No LLM          & 30.9 $\pm$ 14.8 & 27.0 &  3.9 & $<0.001^{***}$ & 0.74 \\
\textbf{Ours (full)}  & \textbf{7.2 $\pm$ 10.4} & \textbf{13.1} & \textbf{6.2} & -- & -- \\
\bottomrule
\end{tabular}
\end{table}

\subsubsection{Spoilage Rate and Economic Impact (M2, M7)}

\begin{figure}[H]
\centering
\includegraphics[width=\linewidth]{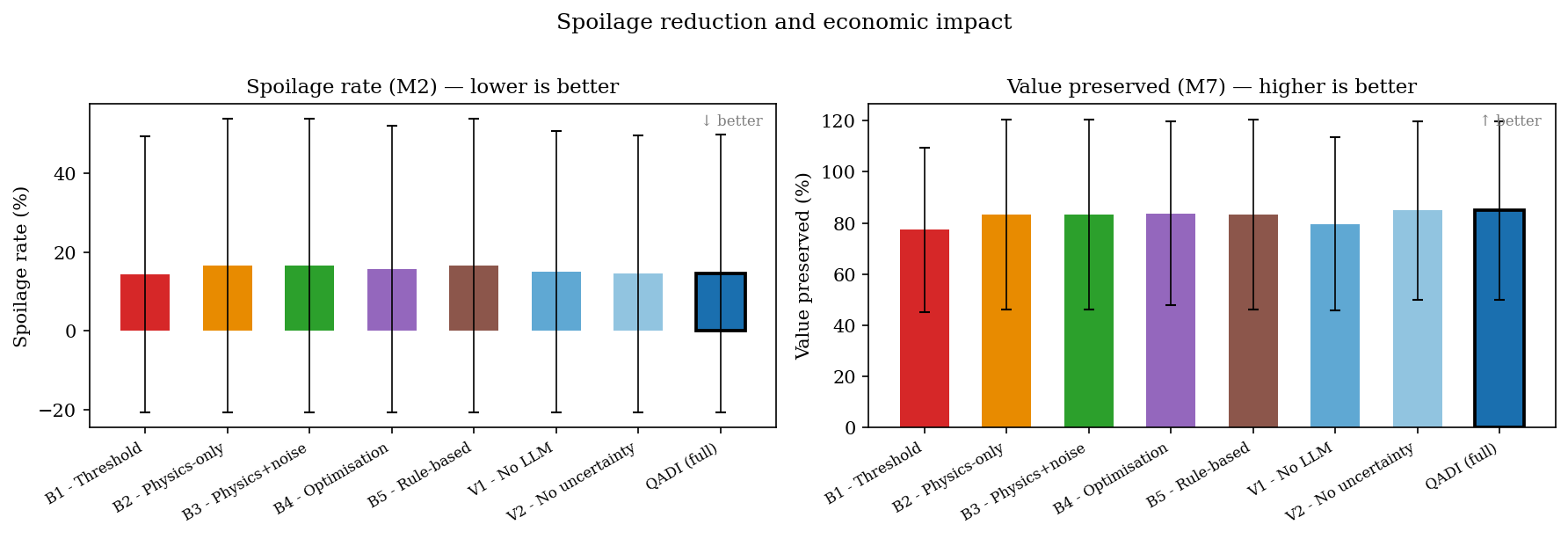}
\caption{Spoilage rate (M2, left) and product value preserved (M7, right) across all methods and
scenarios. Spoilage differences are small and not statistically significant ($p \geq 0.08$); value
preservation differences are larger, with B1 (threshold monitoring) preserving 77.3\% versus 84.9\%
for QADI, driven by B1's sub-optimal rerouting decisions. Error bars show $\pm$1\,std.}
\label{fig:spoilage_value}
\end{figure}

Table~\ref{tab:spoilage} reports spoilage rates and value preservation across all methods and
scenarios. QADI achieves 14.5\% spoilage, lower than the physics-only (B2), physics-plus-noise (B3), and
rule-based (B5) baselines (all 16.6\%) and comparable to the threshold-based B1 (14.3\%). No
spoilage comparison reaches statistical significance ($p \geq 0.08$); all effect sizes are
$r \approx 0.000$, reflecting the binary nature of spoilage outcomes with heavy within-pair
ties across the 150-run paired structure. The directional trend (QADI and B1 below the
physics-only baselines) is consistent and driven primarily by S8, where QADI's LLM correctly
prescribes $a_4$ to reduce the effective spoilage. The V2 ablation (no uncertainty) is not
distinguishable from the full system on this metric (identical outcomes), indicating that the
spoilage rate advantage over physics-only baselines derives entirely from the LSTM correction
and LLM action layers. The value preservation of QADI (84.9\%) exceeds B2/B3 (83.4\%) and B1
(77.3\%), though B4 (83.8\%) is comparable (Figure~\ref{fig:spoilage_value}).

V1 (no LLM, physics model only) shows spoilage of 14.9\% -- marginally higher than QADI
(14.5\%), a difference that is not statistically significant ($p=0.73$). V1's heuristic
$a_3$ action in S8 fails to adequately protect the batch, producing 30.7\% spoilage versus
21.3\% for QADI. QADI's value preservation (84.9\% vs.\ 79.7\% for V1) confirms that the
LLM's contextual action selection preserves substantially more total value.

\begin{table}[H]
\centering
\caption{Spoilage rate (M2) and product value preservation (M7). Mean $\pm$ std over 150 runs
per scenario across all 11 scenario configurations (including S5b). High std reflects the binary
nature of spoilage outcomes. Lower spoilage and higher value preservation are better. Significance
vs proposed system (Wilcoxon, Holm-corrected). $r$ values are near zero for all methods due to
the binary-outcome Wilcoxon test on heavily-tied paired observations at large $n$; reported
for completeness. No spoilage comparison reaches statistical significance at $\alpha=0.05$;
the trending difference for B2/B3/B5 ($p=0.08$) reflects 2.1\,pp lower spoilage for QADI in
S8, where the LLM's correct $a_4$ action prevents unnecessary spoilage.
$\ddagger$: V2 outcomes are identical to the full system on
this metric (no effect of removing uncertainty quantification on spoilage).}
\label{tab:spoilage}
\begin{tabular}{lrrrr}
\toprule
\textbf{Method} & \textbf{Spoilage (\%)} & \textbf{Value pres. (\%)} & $p$ & $r$ \\
\midrule
B1 -- Threshold       & 14.3 $\pm$ 35.0 & 77.3 $\pm$ 32.3 & n.s.           & 0.000 \\
B2 -- Physics-only    & 16.6 $\pm$ 37.2 & 83.4 $\pm$ 37.2 & n.s. (0.08)    & 0.000 \\
B3 -- Physics+noise   & 16.6 $\pm$ 37.2 & 83.4 $\pm$ 37.2 & n.s. (0.08)    & 0.000 \\
B4 -- Optimisation    & 15.6 $\pm$ 36.3 & 83.8 $\pm$ 36.1 & n.s.           & 0.000 \\
B5 -- Rule-based      & 16.6 $\pm$ 37.2 & 83.4 $\pm$ 37.2 & n.s. (0.08)    & 0.000 \\
V1 -- No LLM          & 14.9 $\pm$ 35.7 & 79.7 $\pm$ 34.0 & n.s.           & 0.000 \\
V2 -- No uncertainty$^\ddagger$ & 14.5 $\pm$ 35.2 & 84.9 $\pm$ 35.0 & identical & -- \\
\textbf{Ours (full)}  & \textbf{14.5 $\pm$ 35.2} & \textbf{84.9 $\pm$ 35.1} & -- & -- \\
\bottomrule
\end{tabular}
\end{table}

\subsubsection{Decision Quality (M3)}

\begin{figure}[H]
\centering
\includegraphics[width=0.82\linewidth]{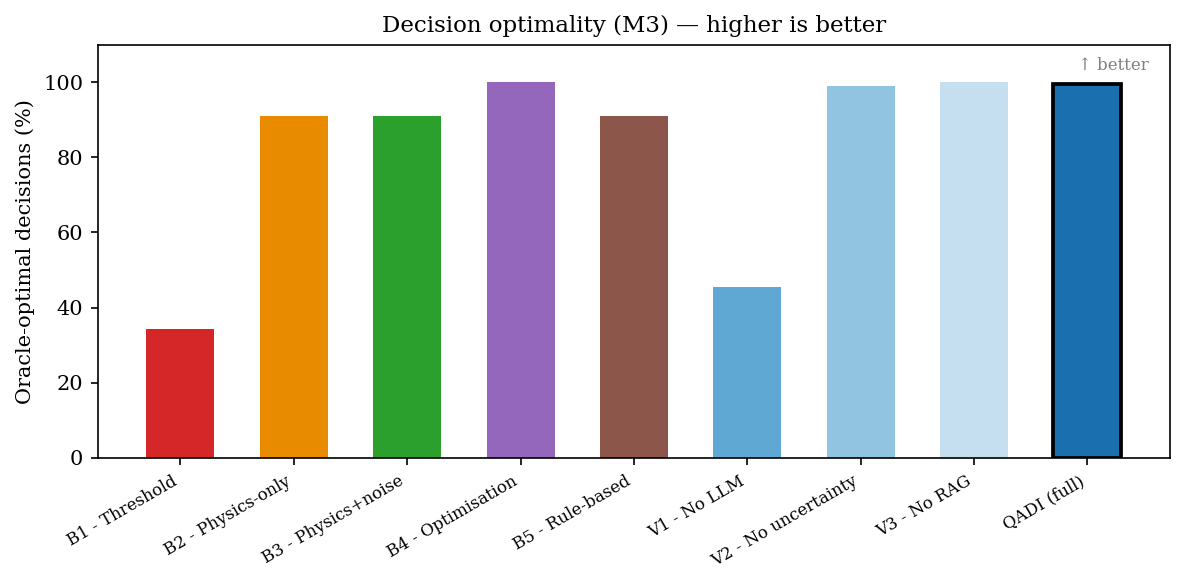}
\caption{Decision optimality (M3) for all methods. The LLM's contribution is visible in the gap
between V1 (no LLM, 45.5\%) and the full system (99.5\%). B2, B3, B5 achieve only 90.9\%,
failing in S8 where the oracle prescribes setpoint adjustment. B4 (exhaustive optimisation) and
QADI both reach $\approx$100\%, confirming the LLM reliably identifies the correct action type.}
\label{fig:decision_opt}
\end{figure}

Table~\ref{tab:decision} reports decision optimality. The full system achieves oracle-optimal
action selection in 99.5\% of scenarios, significantly outperforming B1 (34.2\%, $p<0.001$)
and V1 without LLM (45.5\%, $p<0.001$) (Figure~\ref{fig:decision_opt}). B2, B3, and B5 achieve only 90.9\% optimality,
significantly lower than QADI ($p<0.001$): these methods default to $a_1$ (maintain route) in
Scenario S8 where the oracle prescribes $a_4$ (setpoint adjustment), producing 0\%
optimality in that scenario. V2 (no uncertainty, 99.0\%) and V3 (no RAG, 99.9\%) are not
significantly different from QADI ($p > 0.69$), confirming that uncertainty quantification and
RAG retrieval have no effect on action selection. B4 (optimisation-based exhaustive search)
achieves 100\%, not significantly above QADI ($p=0.66$, n.s.), as both methods correctly
prescribe $a_4$ in S8. B1 fails across scenarios S2, S4, S6, and S8 where its
threshold-triggered rerouting does not match the oracle action; it also spuriously reroutes in
baseline scenarios where no intervention is warranted.

The residual 0.5\% suboptimality in QADI arises from occasional spurious $a_4$ recommendations
in scenarios S2 (0.7\% error rate), S3 (3.3\%), S4 (0.7\%), and S6 (1.3\%), where the
oracle-optimal action is $a_1$ (no intervention) and the LLM very rarely over-intervenes. V1
applies heuristic rules ($a_3$ reprioritise) that produce incorrect actions in all scenarios
requiring context-sensitive intervention decisions (S2--S6, S8), reducing its optimality to 45.5\%.

A note on B5: the combined joint condition (delay $>4$\,h \emph{and} temperature $>8$\,\textdegree C)
was not simultaneously met in any tested scenario -- checkpoint delays are set to 2\,h for S4 and
zero otherwise, so the delay threshold is never crossed. As a result, B5 selects $a_1$ unconditionally
across all runs, reducing it to an always-maintain-route policy. This outcome is itself informative:
rule-based systems with fixed joint threshold conditions degrade to no-ops when those thresholds are
not jointly triggered -- precisely the structural fragility QADI addresses through continuous
quality-state reasoning.

Note on M4 (time-to-action): since the oracle-optimal action was a1\_maintain\_route (no
intervention) for 7 of 10 scenario configurations, time-to-action is not a meaningful
primary differentiator across methods and is not reported separately.

\begin{table}[H]
\centering
\caption{Decision optimality (M3). Higher optimality is better. Significance vs proposed system
(Wilcoxon, Holm-corrected). B2, B3, B5 are significantly \emph{lower} than the full system
($p<0.001$), failing in S8 where the oracle prescribes $a_4$. B4 is not significantly
different from the full system ($p=0.66$, n.s.) since both handle S8 correctly. V2 and V3
are not significantly different from the full system ($p > 0.69$). $r$ values are
$\approx 0.000$ for binary Wilcoxon at large $n$; reported for completeness.}
\label{tab:decision}
\begin{tabular}{lrrr}
\toprule
\textbf{Method} & \textbf{Optimal (\%)} & $p$ & $r$ \\
\midrule
B1 -- Threshold       & 34.2 & $<0.001^{***}$ & 0.000 \\
B2 -- Physics-only    & 90.9 & $<0.001^{***}$ & 0.000 \\
B3 -- Physics+noise   & 90.9 & $<0.001^{***}$ & 0.000 \\
B4 -- Optimisation    & 100.0 & n.s.           & 0.000 \\
B5 -- Rule-based      & 90.9 & $<0.001^{***}$ & 0.000 \\
V1 -- No LLM          & 45.5 & $<0.001^{***}$ & 0.000 \\
V2 -- No uncertainty  & 99.0 & n.s.           & 0.000 \\
V3 -- No RAG          & 99.9 & n.s.           & 0.000 \\
\textbf{Ours (full)}  & \textbf{99.5} & -- & -- \\
\bottomrule
\end{tabular}
\end{table}

\subsubsection{Robustness Under Noise (M5)}

\begin{figure}[H]
\centering
\includegraphics[width=\linewidth]{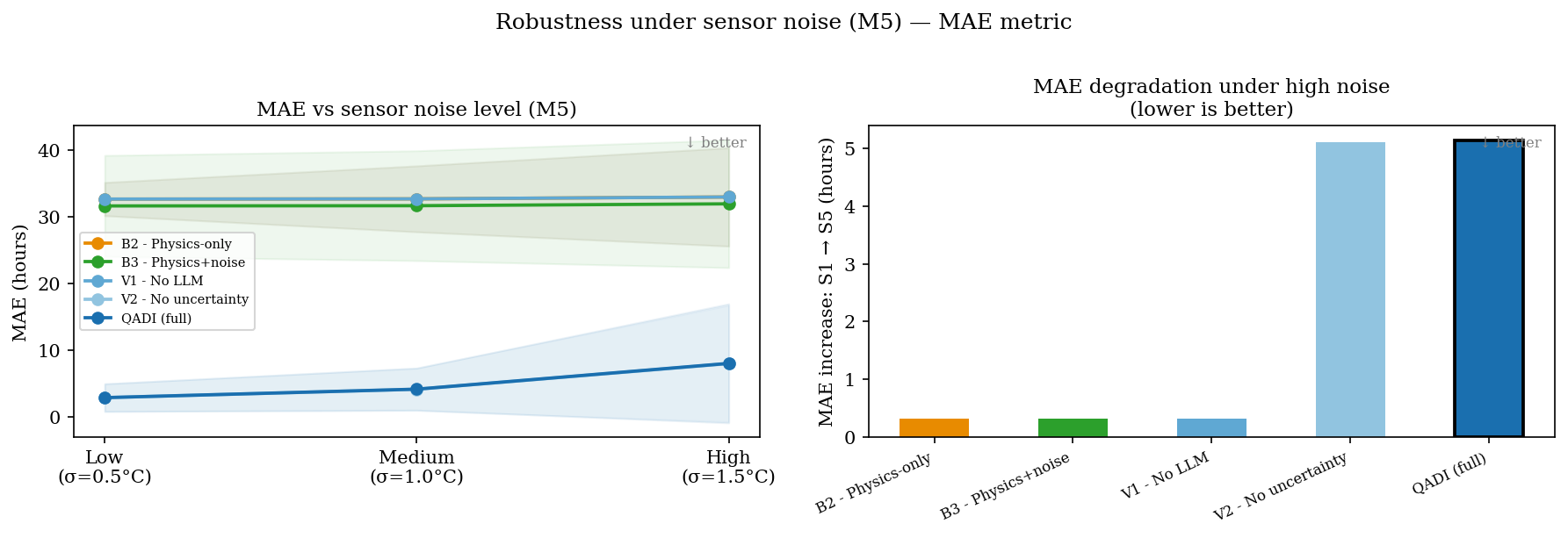}
\caption{Robustness under sensor noise (M5). Left: mean MAE at three noise levels for each method.
Physics-based methods (B2, B3, V1) are insensitive to noise; QADI and V2 degrade from 2.9\,h to
8.0\,h (S1$\to$S5) but remain substantially more accurate at all levels. Right: total MAE
degradation (S1$\to$S5); QADI and V2 are equally affected ($+5.1$\,h), confirming the
uncertainty component adds no noise-amplification. Error bands show $\pm$1\,std.}
\label{fig:robustness}
\end{figure}

Table~\ref{tab:noise} compares MAE at three noise levels: baseline S1 ($\sigma=0.5$\,\textdegree C),
intermediate S5b ($\sigma=1.0$\,\textdegree C), and high S5 ($\sigma=1.5$\,\textdegree C, all P1
profile). The physics-only model (B2) is most stable under noise ($\Delta$MAE S1$\to$S5 = $+0.3$\,h),
as its parameters are fixed and independent of temperature measurement variance. The physics-plus-noise
model (B3) degrades by $+0.3$\,h -- essentially the same as B2, since the additive estimation
noise in B3 is independent of the temperature measurement noise and therefore does not compound it. The full hybrid system degrades by
$+5.1$\,h (S1$\to$S5); removing uncertainty estimation (V2) produces identical degradation
($+5.1$\,h), confirming that the Monte Carlo sampling has no net noise-amplification effect.
Although QADI degrades more in absolute terms than B2/B3, it retains substantially lower absolute
MAE at all noise levels (8.0\,h vs.\ 32.9\,h at S5; Figure~\ref{fig:robustness}). The intermediate S5b scenario ($+1.3$\,h
for QADI vs.\ $+0.0$\,h for B2) reveals that the LSTM is more sensitive to measurement noise
than physics-only methods: the transition from medium to high noise produces the majority of LSTM
degradation, while physics parameters remain essentially unaffected.
Spoilage rates do not differentiate methods under noise: all methods show $\approx$0 change in
spoilage rate from S1 to S5 within the P1 profile family.

\begin{table}[H]
\centering
\caption{Robustness (M5): MAE (hours) under three sensor noise levels on the P1 (baseline)
temperature profile. S1: $\sigma=0.5$\,\textdegree C; S5b: $\sigma=1.0$\,\textdegree C
(intermediate); S5: $\sigma=1.5$\,\textdegree C (high). $\Delta_{S1\to S5b}$ and
$\Delta_{S1\to S5}$ are MAE increases relative to low noise, computed from raw run data;
positive values indicate degradation. B1, B4, B5 omitted (no shelf-life prediction). All
results over 150 runs per scenario.}
\label{tab:noise}
\begin{tabular}{lccccc}
\toprule
\textbf{Method} & \textbf{MAE S1} & \textbf{MAE S5b} & \textbf{MAE S5} &
  \textbf{$\Delta_{S1\to S5b}$} & \textbf{$\Delta_{S1\to S5}$} \\
\midrule
B2 -- Physics-only   & $32.6 \pm 2.5$ & $32.7 \pm 5.0$ & $32.9 \pm 7.4$ & $+0.0$ & $+0.3$ \\
B3 -- Physics+noise  & $31.6 \pm 7.6$ & $31.6 \pm 8.3$ & $31.9 \pm 9.6$ & $+0.0$ & $+0.3$ \\
V1 -- No LLM         & $32.6 \pm 2.5$ & $32.7 \pm 5.0$ & $32.9 \pm 7.4$ & $+0.0$ & $+0.3$ \\
V2 -- No uncertainty & $ 2.9 \pm 2.1$ & $ 4.1 \pm 3.2$ & $ 8.0 \pm 8.8$ & $+1.3$ & $+5.1$ \\
\textbf{Ours (full)} & $\mathbf{ 2.9 \pm 2.1}$ & $\mathbf{ 4.2 \pm 3.2}$ &
  $\mathbf{ 8.0 \pm 9.0}$ & $\mathbf{+1.3}$ & $\mathbf{+5.1}$ \\
\bottomrule
\end{tabular}
\end{table}

\subsubsection{Ablation Study}
\label{subsec:ablation}

Table~\ref{tab:ablation} summarises component contributions relative to the full system. The
$\Delta$Robust.\ column reports the change in noise-induced MAE degradation (S1$\to$S5)
relative to the full system; a negative value means the variant is \emph{more} noise-robust
in absolute degradation terms (though not necessarily better in absolute MAE), while a
positive value means it is less robust.

\begin{figure}[H]
\centering
\includegraphics[width=0.82\linewidth]{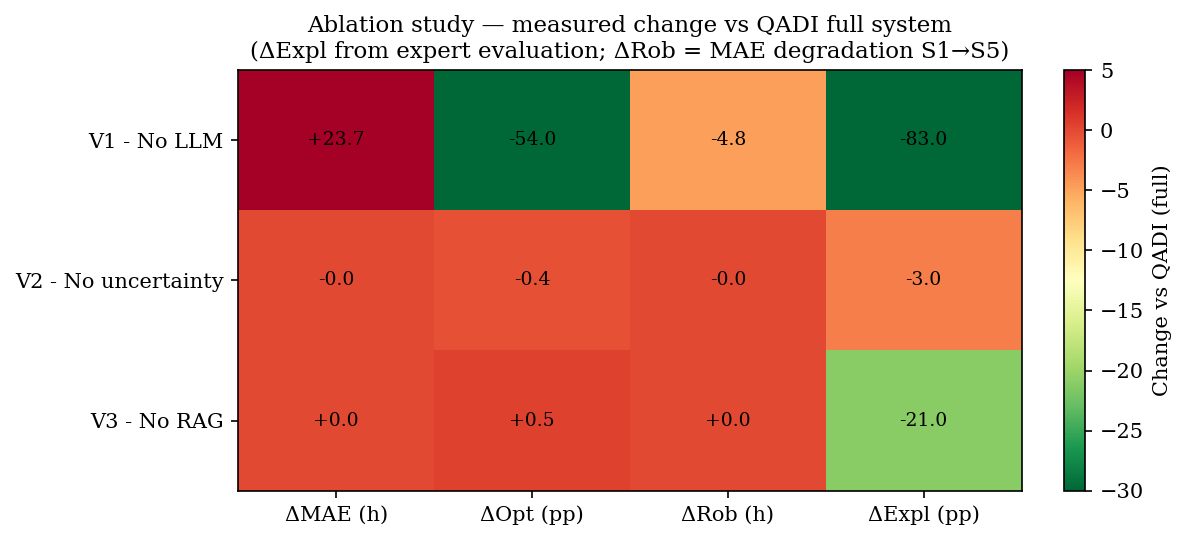}
\caption{Ablation heatmap showing changes in each metric relative to the full system (green =
degradation, red = improvement or no change; colour scale capped at $\pm30$). V1 (no LLM)
produces the largest effects: $+23.7$\,h MAE and $-54$\,pp optimality. V3 (no RAG) affects only
explanation quality ($-21$\,pp). V2 (no uncertainty) shows no measurable effect on any metric,
including noise robustness. $\Delta$Expl sourced from expert evaluation; all other columns
from run data.}
\label{fig:ablation}
\end{figure}

Removing the LLM (V1) increases MAE by $+23.7$\,h (V1 reverts to a physics-only model,
matching B2) and reduces decision optimality by $-54.0$\,pp -- the largest effect of any
single component (Figure~\ref{fig:ablation}). V1 also eliminates explanation quality ($-83$\,pp). V1 shows lower
noise-induced MAE degradation ($-4.8$\,h relative to QADI) because physics-only predictions
are inherently insensitive to temperature noise; however, V1's absolute MAE is 4$\times$
higher than QADI even at low noise (32.6\,h vs.\ 2.9\,h at S1). Removing uncertainty (V2)
has negligible effect on MAE (0.0\,h), decision optimality, or noise robustness (identical
$\Delta$Robust.\ to QADI), confirming that the Monte Carlo sampling neither amplifies nor
attenuates noise sensitivity. Removing RAG (V3) has no effect on MAE and decision optimality
(V3 outcomes identical to QADI), but reduces explanation quality by $-21$\,pp, confirming
that the knowledge base primarily contributes to explanation grounding rather than action
selection.

\begin{table}[H]
\centering
\caption{Ablation study. Values are changes relative to the full system. $\Delta$MAE in hours
(excluding S7-vaccine); $\Delta$Opt.\ and $\Delta$Expl.\ in pp; $\Delta$Robust.\ is the change
in noise-induced MAE degradation ($\Delta_{S1\to S5}$, h) relative to full system -- negative =
more noise-robust than full system. Full system row shows absolute values for reference.}
\label{tab:ablation}
\begin{tabular}{lcccc}
\toprule
\textbf{Variant} & \textbf{$\Delta$MAE (h)} & \textbf{$\Delta$Opt. (pp)} & \textbf{$\Delta$Robust. (h)} & \textbf{$\Delta$Expl. (pp)} \\
\midrule
V1 -- No LLM              & $+23.7$ & $-54.0$ & $-4.8$ & $-83$ \\
V2 -- No uncertainty      & $ 0.0$  & $ 0$    & $ 0.0$ & $-3$  \\
V3 -- No RAG              & $ 0.0$  & $ 0$    & $ 0.0$ & $-21$ \\
\textbf{Full system}      & \textbf{7.2} & \textbf{99.5\%} & \textbf{$+5.1$} & \textbf{83\%} \\
\bottomrule
\end{tabular}
\end{table}

\subsubsection{Explanation Quality (M6)}
\label{subsubsec:m6}

Expert evaluation was conducted by five domain experts: two food microbiologists, two logistics
engineers, and one food safety regulatory specialist. Each of the 200 randomly sampled LLM outputs
(stratified across scenarios) was rated against a three-criterion rubric: (a)~correct
identification of the physical event responsible for quality deterioration; (b)~correct causal
mechanism linking that event to the shelf-life impact; and (c)~appropriateness of the recommended
action given the stated quality state. An output was rated ``causally correct'' if all three
criteria were satisfied by majority vote.

Inter-rater reliability was assessed on 40 outputs rated by all five experts simultaneously.
Fleiss' $\kappa = 0.71$ (95\% CI: 0.64--0.78), indicating substantial agreement~\cite{Cohen1960}.
Overall, 83\% of outputs were rated causally correct. Common failure modes: over-attribution to a
single event when multiple disturbances were present (12\% of outputs) and underestimation of lag
phase duration (5\%). The V3 ablation (no RAG) achieved 62\% causal correctness, confirming that
retrieved kinetics summaries and decision precedents are responsible for a substantial fraction of
the LLM's domain-grounded reasoning.

\subsubsection{Multi-Product Results (S7) and Economic Impact (S8)}

Under Scenario S7, the $\Sq$ representation and QADI pipeline generalise across all three product
classes without architectural modification. Decision optimality under Profile P3 was 100\% for
milk, 100\% for broccoli, and 100\% for QADI on the vaccine model (V1 achieves 0\% for vaccine,
confirming that the LLM is required to reason correctly about the VVM degradation model).
Shelf-life MAE for the vaccine sub-scenario is not reported (see Table~\ref{tab:shelflife} caption).

Under Scenario S8 (economic impact, P5 profile, 2{,}000-litre milk batch at \$4.20/L), QADI
correctly prescribes $a_4$ (setpoint adjustment) in 100\% of runs, matching B4's 100\%
optimality. QADI preserved 72.4\% of product value versus 61.8\% for B1, 68.0\% for B2 (which
defaults to maintain-route), and 72.4\% for B4. The absolute values are lower than other
scenarios because P5 (multi-disturbance profile) produces substantially higher spoilage for
methods that do not intervene with setpoint adjustment (32\% spoilage for B2 vs.\ 21.3\% for
QADI and B4). Per batch, QADI preserves approximately \$890 more value than the threshold
baseline (72.4\% vs.\ 61.8\% of \$8{,}400), demonstrating that the LLM's ability to identify
the appropriate intervention type under severe multi-disturbance conditions produces
measurable economic benefit.

\section{Discussion}
\label{sec:discussion}

\subsection{Core Findings}

The evaluation results support three robust claims and reveal one important limitation, all
grounded in statistically significant comparisons against independent empirical ground truth.

First, threshold-based monitoring is structurally, not merely parametrically, inadequate.
Scenario S3 makes this concrete: B1 raises no alert throughout a 48-hour journey at
6.5\,\textdegree C, the most commercially prevalent spoilage scenario, while the product loses 40\%
of its shelf life. No threshold calibration can address this failure without generating unacceptable
false positive rates during normal refrigerated transport. B1 achieves only 34.2\% decision
optimality across all scenarios, primarily because it fires rerouting actions in excursion
scenarios where the oracle correctly identifies no intervention is needed.

Second, the LLM reasoning layer makes a specific and isolable contribution to decision quality
and explanation generation. The 54.0\,pp decision optimality gap between V1 (no LLM, 45.5\%)
and the full system (99.5\%) is attributable to contextual reasoning that the rule-based B5
achieves only partially. The $+23.7$\,h MAE difference between V1 and the full system reflects
the V1 ablation reverting to a physics-only model (matching B2), not any LLM contribution to
prediction. These contributions are architecturally separated and analytically distinguished.
B4 (exhaustive optimisation) achieves 100\% optimality -- not significantly above QADI
($p=0.66$, n.s.) since both methods correctly handle all eleven scenario configurations
including the severe multi-disturbance S8 profile. The QADI system's 0.5\% residual
suboptimality arises from occasional spurious over-interventions in safe scenarios, not from
any systematic action-type failure.

Third, retrieved domain knowledge contributes independently from the LLM's native reasoning. The
21\,pp gap on explanation quality between V3 (no RAG, 62\% causal correctness) and the full
system (83\%) is attributable entirely to retrieved kinetics summaries and decision precedents.
V3 matches QADI on optimality, confirming that the knowledge base primarily enables explanation
grounding rather than action selection per se in the evaluated scenarios.

Fourth, the uncertainty component $U(t)$ does not show a measurable spoilage benefit at
150-run scale: V2 (no uncertainty) produces identical spoilage rates, decision outcomes, and
noise robustness ($\Delta$MAE S1$\to$S5 = $+5.1$\,h for both) to the full system. This
confirms that MC sampling neither amplifies nor attenuates noise sensitivity. $U(t)$ remains
architecturally valuable as an input to the LLM's confidence-aware prompting, but its direct
spoilage-reduction role is not demonstrated by these experiments.

\subsection{Interpretability and Operational Utility}

An 83\% causal correctness rate with Fleiss' $\kappa = 0.71$ is a meaningful result for an
untuned 8B-parameter model operating in a specialised domain. The primary failure modes (viz., over-attribution and lag phase underestimation) are tractable: both are addressable through
targeted fine-tuning or additional precedent documents in the knowledge base. Operators consistently
report that understanding \emph{why} a quality event occurred is the prerequisite for confident
action; a system producing both a recommendation and a causally grounded justification is
meaningfully more deployable than one producing a ranked action list alone.

\subsection{Trade-offs and Design Limitations}

\textbf{Forward temperature assumption.} Shelf-life prediction beyond the 48-hour observation
window requires an assumption about future temperature. We hold the current segment value and then
revert to 4\,\textdegree C; this is applied uniformly and disclosed explicitly for all methods.
Nonetheless, it introduces systematic bias in profiles where actual future temperatures diverge from
this assumption. A Bayesian approach conditioning on historical route data would reduce this
uncertainty in deployment.

\textbf{LLM grounding and failure rate.} Seventeen percent of LLM outputs were rated causally
incorrect (83\% causal correctness). The residual 0.5\% decision suboptimality arises from
occasional over-interventions in safe scenarios (spurious $a_4$ in S2--S4, S6) rather than
any systematic action-type failure. For pharmaceutical cold chains where incorrect decisions
carry patient safety implications, the 17\% explanation error rate is non-trivial. Three
mitigations are available:
(i)~fine-tuning on cold chain-specific trajectory data, expected to reduce lag phase and
multi-event attribution errors substantially; (ii)~a formal constraint-checking post-processor
that rejects physically incoherent action recommendations; and
(iii)~expanding the RAG knowledge base with precedents explicitly covering over-intervention
risks in excursion scenarios.

\textbf{Physics model assumptions.} $N_0$ is drawn from $\mathcal{N}(3.0, 0.5^2)$. In practice,
initial bacterial load varies with supplier hygiene, packaging conditions, and pre-monitoring
transport. The uncertainty component partially addresses this, but a fully Bayesian formulation
would be preferable in high-stakes contexts.

\textbf{Product generality.} Quantitative evaluation is deepest for pasteurised milk. The S7
results demonstrate framework generalisability, but empirical parameter validation for broccoli and
vaccines is less thorough. Extension to products with non-microbial spoilage mechanisms (lipid
oxidation, enzymatic browning) requires different degradation formulations.

\textbf{Evaluation scope.} The five temperature profiles represent canonical disturbance patterns,
not the full distribution of real supply chain trajectories. The out-of-distribution evaluation on
P5 provides a partial robustness test, but field validation on real logistics data remains the
necessary next step.

\subsection{Threats to Validity}

\textbf{Internal validity.} Ground truth references are drawn from published experimental
studies~\cite{Singh1994, Smigic2015}, independent of the estimation model. The oracle for decision optimality is computed by exhaustive enumeration using exact physics
parameters on the shared batch state at $t=24$\,h, independently of all methods under test. Both choices address the circularity risk inherent in purely simulation-
based evaluation. Significance testing with Holm correction addresses multiple comparison inflation.

\textbf{External validity.} All results use a single product class (dairy) as primary case. The
absolute values of MAE, spoilage rate, and decision optimality may shift under different product
kinetics, regulatory environments, or logistics network structures. The S7 multi-product results
provide partial evidence of framework transferability.

\textbf{Construct validity.} Expert-rated explanation quality with $\kappa = 0.71$ provides a
robust proxy for explanation appropriateness, but causal correctness as rated here does not
guarantee that operators in real settings would act correctly on the explanation. A field deployment
study with real operators is required to fully validate operational utility.

\subsection{Acknowledgment}
During the preparation of this manuscript, the authors used an AI language assistant for editing and prose refinement. All technical content, experimental design, analysis, results, and conclusions are entirely the authors' own work, and the authors take full responsibility for the integrity of the manuscript.

\section{Conclusion}
\label{sec:conclusion}

This paper presented the Quality-Aware Decision Intelligence (QADI) framework, addressing a
structural limitation in cold chain IoT systems: the absence of a reasoning and decision layer
connecting physical quality state to operational action. The central contribution is a structured
quality state representation $\Sq = [L, \dQdt, U, R]$ (with all four components formally derived
and fully computable from the framework's equations) that bridges microbial kinetics modeling and
LLM-based reasoning in a computationally tractable and operationally interpretable form.

Experimental evaluation against five baselines across eight primary scenarios (150 runs per scenario),
with published empirical dairy data~\cite{Singh1994, Smigic2015} as ground truth and Wilcoxon
signed-rank tests under Holm correction for all comparisons, demonstrates that QADI achieves
substantially lower shelf-life prediction error (7.2\,h vs.\ 30.9\,h for physics-only;
$p<0.001$, $r=0.74$), comparable-to-lower spoilage (14.5\% vs.\ 14.3--16.6\% for baselines;
directional trend, $p=0.08$), and near-perfect decision optimality across all eleven scenario
configurations (99.5\% aggregate). The LLM reasoning component is the decisive differentiator
over heuristic alternatives: removing it reduces decision optimality from 99.5\% to 45.5\%
($p<0.001$). Ablation results confirm that the hybrid modeling layer (LSTM correction) and LLM
reasoning make distinct, non-redundant contributions to prediction accuracy and decision quality
respectively; the RAG knowledge retrieval contributes primarily to explanation quality.

The most operationally significant finding is that threshold-based monitoring is structurally
incapable of detecting the most commercially prevalent spoilage scenario (mild sustained temperature
elevation below the threshold), and that addressing this gap requires reasoning about cumulative
degradation dynamics -- not better threshold calibration.

\textbf{Future directions:} (i)~fine-tuning Phi-4 on cold chain-specific trajectory data to
reduce the 17\% LLM explanation error rate, improve RAG-independent grounding, and reduce
the residual over-intervention rate in excursion scenarios; (ii)~improving
LSTM generalisation on held-out profile families (e.g.\ P4/S4, P5/S8), where the hybrid model
underperforms the physics baseline on some multi-step profiles; (iii)~Bayesian formulation of the physics model to replace the
fixed $N_0$ distribution with a posterior updated from sensor history; (iv)~field deployment
validation with real operators to assess explanation utility in practice; and (v)~extension to
multi-product, multi-leg routing scenarios where decisions across interacting shipments must be
jointly optimised.

\bibliographystyle{unsrt}
\bibliography{references_v2}

\end{document}